\pdfoutput=1
\documentclass[10pt,twocolumn,letterpaper]{article}
\usepackage[T1]{fontenc}
\usepackage[pagenumbers]{cvpr}
\usepackage{graphicx}
\usepackage{booktabs}
\usepackage{amsmath,amssymb}
\usepackage{xcolor}
\usepackage{multirow}
\usepackage{algorithm}
\usepackage{algpseudocode}
\usepackage[numbers,sort&compress]{natbib}
\definecolor{cvprblue}{rgb}{0.21,0.49,0.74}
\usepackage[breaklinks,colorlinks,allcolors=cvprblue]{hyperref}

\newcommand{\ablInterSign}{\ensuremath{-}0.69}

\newcommand{\ablRmRec}{1.01}

\newcommand{\ablRmRecImgs}{248}

\newcommand{\adaptGainMax}{3.1}
\newcommand{\adaptGainMin}{2.5}
\newcommand{\aseedSdAll}{0.01}
\newcommand{\aseedSdDoc}{0.50}

\newcommand{\bufAdKFourMax}{0.26}
\newcommand{\bufAdKFourMin}{\ensuremath{-}0.62}

\newcommand{\bufAdKOursMax}{0.45}
\newcommand{\bufAdKOursMin}{\ensuremath{-}0.04}
\newcommand{\bufGain}{0.43}

\newcommand{\bufKdeepMean}{2.1}

\newcommand{\bufKdeepTime}{20}
\newcommand{\bufKmax}{8}
\newcommand{\bufKoneMean}{81}

\newcommand{\bufTruncRuns}{18}

\newcommand{\corpusTotal}{102{,}274}

\newcommand{\cpuEncS}{0.9}
\newcommand{\cpuEncVsJxl}{3.5}

\newcommand{\cvsKodak}{12.9}

\newcommand{\cvsRs}{5.1}

\newcommand{\finFnlicOverSmac}{2.7}

\newcommand{\finJxlDecEightMs}{54.9}

\newcommand{\finJxlDecOneMs}{387}

\newcommand{\finLmac}{143.7}

\newcommand{\finLpar}{144}
\newcommand{\finLvsCallic}{0.8}

\newcommand{\finMmac}{48.1}

\newcommand{\finRatioCallicL}{4}
\newcommand{\finRatioCallicM}{12}
\newcommand{\finRatioCallicS}{51}

\newcommand{\finRegressN}{6,000}
\newcommand{\finSFnlicMax}{21.3}
\newcommand{\finSFnlicMin}{1.1}

\newcommand{\finScpuVsCallicMax}{73}
\newcommand{\finScpuVsCallicMin}{47}
\newcommand{\finSdecEightMs}{55.5}
\newcommand{\finSdecEightPlainMs}{52.5}

\newcommand{\finSdecOneMs}{414}

\newcommand{\finSencS}{2.15}

\newcommand{\finSmac}{11.3}

\newcommand{\finSnoVsVsJxl}{7.7}
\newcommand{\finSoverCallic}{2.9}

\newcommand{\finSpar}{11.5}

\newcommand{\finSvsJxl}{8.4}

\newcommand{\fnlicEncH}{3.4}

\newcommand{\hClsJls}{3.052}

\newcommand{\hiVsGain}{6.3}

\newcommand{\ledAdaMax}{10.6}
\newcommand{\ledAdaMaxSet}{Doc}
\newcommand{\ledCross}{35}

\newcommand{\ledFeatGumus}{0.7}

\newcommand{\ledPixGumus}{2.3}

\newcommand{\ledStateMax}{4.0}
\newcommand{\ledStateMaxSet}{RS19}

\newcommand{\looMax}{3.36}
\newcommand{\looMin}{0.44}
\newcommand{\looRefit}{0.45}

\newcommand{\macAribM}{136}
\newcommand{\macCallic}{580}

\newcommand{\nSeven}{250}

\newcommand{\noaInterSign}{+0.06}

\newcommand{\parGumusNominal}{59}
\newcommand{\parGumusStored}{64}

\newcommand{\parLlictiK}{197}

\newcommand{\pixShareDivClic}{89}

\newcommand{\realOverIdeal}{0.06}

\newcommand{\tabNoGrad}{0.60}

\newcommand{\tabTbpeVsNonrec}{1.8}

\newcommand{\tabTbpttEager}{17}
\newcommand{\tabTbpttSame}{0.58}

\newcommand{\tabTbpttSlow}{52}

\newcommand{\tseedSdAll}{0.14}
\newcommand{\tseedSdDoc}{0.76}

\newcommand{\vsAllGain}{0.79}

\newcommand{\vsAlwaysGain}{0.77}

\newcommand{\vsCoarseOnLoveDA}{12}

\newcommand{\vsDecRMax}{0.18}

\newcommand{\vsKodakGain}{12.2}

\newcommand{\vsOnLoveda}{16}

\newcommand{\vsOtherCostMax}{0.11}

\newcommand{\xdevCpuN}{5}

\newcommand{\xdevGpuN}{5}

\newcommand{\xdevVsGpuN}{3}

\newcommand{\sTdStateLoAll}{7.0}
\newcommand{\sTdStateHiAll}{9.1}
\newcommand{\sTdBitsLoAll}{5.3}
\newcommand{\sTdBitsHiAll}{15.2}
\newcommand{\sZsOneVsTable}{20}
\newcommand{\sZsTwoVsTable}{68}
\newcommand{\sTbfVsNr}{1.0}
\newcommand{\sTbfUpd}{112}

\newcommand{\sNopgoMin}{1.5}
\newcommand{\sNopgoMax}{2.9}

\newcommand{\sSeqDepthMean}{60}

\newcommand{\sDepAdBest}{1.36}

\newcommand{\sDepAdUFourFb}{2.79}
\newcommand{\sVtwoGainSEightZeroZero}{0.48}
\newcommand{\sIntCostTwenty}{0.53}

\newcommand{\sWideExtraMacs}{54}
\newcommand{\sIsaAvxOne}{1.6}
\newcommand{\sRssS}{92}
\newcommand{\sRssJxl}{15}

\newcommand{\sMixAll}{0.67}
\newcommand{\sMixDecEight}{46}
\newcommand{\sWebList}{400}
\newcommand{\sCamKept}{398}
\newcommand{\sMproRatioEight}{1.42}

\expandafter\def\csname supplab@sec:impl\endcsname{1}
\expandafter\def\csname supplab@sec:impl@cref\endcsname{[section][1][]1}
\expandafter\def\csname supplab@sec:feat\endcsname{1.1}
\expandafter\def\csname supplab@sec:feat@cref\endcsname{[subsection][1][1]1.1}
\expandafter\def\csname supplab@sec:net\endcsname{1.2}
\expandafter\def\csname supplab@sec:net@cref\endcsname{[subsection][2][1]1.2}
\expandafter\def\csname supplab@sec:train\endcsname{1.3}
\expandafter\def\csname supplab@sec:train@cref\endcsname{[subsection][3][1]1.3}
\expandafter\def\csname supplab@tab:feat\endcsname{S1}
\expandafter\def\csname supplab@tab:feat@cref\endcsname{[table][1][]S1}
\expandafter\def\csname supplab@tab:net\endcsname{S2}
\expandafter\def\csname supplab@tab:net@cref\endcsname{[table][2][]S2}
\expandafter\def\csname supplab@sec:adapt\endcsname{1.4}
\expandafter\def\csname supplab@sec:adapt@cref\endcsname{[subsection][4][1]1.4}
\expandafter\def\csname supplab@alg:enc\endcsname{1}
\expandafter\def\csname supplab@alg:enc@cref\endcsname{[algorithm][1][]1}
\expandafter\def\csname supplab@tab:sideinfo\endcsname{S3}
\expandafter\def\csname supplab@tab:sideinfo@cref\endcsname{[table][3][]S3}
\expandafter\def\csname supplab@sec:stream\endcsname{1.5}
\expandafter\def\csname supplab@sec:stream@cref\endcsname{[subsection][5][1]1.5}
\expandafter\def\csname supplab@sec:p20\endcsname{1.5}
\expandafter\def\csname supplab@sec:p20@cref\endcsname{[subsection][5][1]1.5}
\expandafter\def\csname supplab@sec:vs_supp\endcsname{1.6}
\expandafter\def\csname supplab@sec:vs_supp@cref\endcsname{[subsection][6][1]1.6}
\expandafter\def\csname supplab@tab:t16\endcsname{S4}
\expandafter\def\csname supplab@tab:t16@cref\endcsname{[table][4][]S4}
\expandafter\def\csname supplab@tab:vs\endcsname{S5}
\expandafter\def\csname supplab@tab:vs@cref\endcsname{[table][5][]S5}
\expandafter\def\csname supplab@sec:decoder\endcsname{1.7}
\expandafter\def\csname supplab@sec:decoder@cref\endcsname{[subsection][7][1]1.7}
\expandafter\def\csname supplab@sec:gpudec\endcsname{1.7}
\expandafter\def\csname supplab@sec:gpudec@cref\endcsname{[subsection][7][1]1.7}
\expandafter\def\csname supplab@sec:determ\endcsname{1.8}
\expandafter\def\csname supplab@sec:determ@cref\endcsname{[subsection][8][1]1.8}
\expandafter\def\csname supplab@tab:determ\endcsname{S6}
\expandafter\def\csname supplab@tab:determ@cref\endcsname{[table][6][]S6}
\expandafter\def\csname supplab@sec:data\endcsname{2}
\expandafter\def\csname supplab@sec:data@cref\endcsname{[section][2][]2}
\expandafter\def\csname supplab@sec:corpus\endcsname{2.1}
\expandafter\def\csname supplab@sec:corpus@cref\endcsname{[subsection][1][2]2.1}
\expandafter\def\csname supplab@tab:corpus\endcsname{S7}
\expandafter\def\csname supplab@tab:corpus@cref\endcsname{[table][7][]S7}
\expandafter\def\csname supplab@tab:evalset\endcsname{S8}
\expandafter\def\csname supplab@tab:evalset@cref\endcsname{[table][8][]S8}
\expandafter\def\csname supplab@sec:timing\endcsname{2.3}
\expandafter\def\csname supplab@sec:timing@cref\endcsname{[subsection][3][2]2.3}
\expandafter\def\csname supplab@sec:extruns_supp\endcsname{2.4}
\expandafter\def\csname supplab@sec:extruns_supp@cref\endcsname{[subsection][4][2]2.4}
\expandafter\def\csname supplab@fig:width\endcsname{S1}
\expandafter\def\csname supplab@fig:width@cref\endcsname{[figure][1][]S1}
\expandafter\def\csname supplab@sec:res\endcsname{3}
\expandafter\def\csname supplab@sec:res@cref\endcsname{[section][3][]3}
\expandafter\def\csname supplab@sec:encab\endcsname{3.2}
\expandafter\def\csname supplab@sec:encab@cref\endcsname{[subsection][2][3]3.2}
\expandafter\def\csname supplab@fig:stepcurve\endcsname{S2}
\expandafter\def\csname supplab@fig:stepcurve@cref\endcsname{[figure][2][]S2}
\expandafter\def\csname supplab@sec:cpugpu\endcsname{3.3}
\expandafter\def\csname supplab@sec:cpugpu@cref\endcsname{[subsection][3][3]3.3}
\expandafter\def\csname supplab@tab:extruns\endcsname{S9}
\expandafter\def\csname supplab@tab:extruns@cref\endcsname{[table][9][]S9}
\expandafter\def\csname supplab@sec:speed\endcsname{3.4}
\expandafter\def\csname supplab@sec:speed@cref\endcsname{[subsection][4][3]3.4}
\expandafter\def\csname supplab@sec:isa\endcsname{3.4}
\expandafter\def\csname supplab@sec:isa@cref\endcsname{[subsection][4][3]3.4}
\expandafter\def\csname supplab@tab:lit\endcsname{S10}
\expandafter\def\csname supplab@tab:lit@cref\endcsname{[table][10][]S10}
\expandafter\def\csname supplab@sec:xdev\endcsname{3.5}
\expandafter\def\csname supplab@sec:xdev@cref\endcsname{[subsection][5][3]3.5}
\expandafter\def\csname supplab@tab:main_all\endcsname{S11}
\expandafter\def\csname supplab@tab:main_all@cref\endcsname{[table][11][]S11}
\expandafter\def\csname supplab@tab:encab\endcsname{S12}
\expandafter\def\csname supplab@tab:encab@cref\endcsname{[table][12][]S12}
\expandafter\def\csname supplab@tab:cpugpu\endcsname{S13}
\expandafter\def\csname supplab@tab:cpugpu@cref\endcsname{[table][13][]S13}
\expandafter\def\csname supplab@sec:classic\endcsname{3.6}
\expandafter\def\csname supplab@sec:classic@cref\endcsname{[subsection][6][3]3.6}
\expandafter\def\csname supplab@tab:dectime\endcsname{S14}
\expandafter\def\csname supplab@tab:dectime@cref\endcsname{[table][14][]S14}
\expandafter\def\csname supplab@tab:t21\endcsname{S15}
\expandafter\def\csname supplab@tab:t21@cref\endcsname{[table][15][]S15}
\expandafter\def\csname supplab@sec:bam\endcsname{3.7}
\expandafter\def\csname supplab@sec:bam@cref\endcsname{[subsection][7][3]3.7}
\expandafter\def\csname supplab@tab:regress\endcsname{S16}
\expandafter\def\csname supplab@tab:regress@cref\endcsname{[table][16][]S16}
\expandafter\def\csname supplab@tab:isa\endcsname{S17}
\expandafter\def\csname supplab@tab:isa@cref\endcsname{[table][17][]S17}
\expandafter\def\csname supplab@tab:gpu\endcsname{S18}
\expandafter\def\csname supplab@tab:gpu@cref\endcsname{[table][18][]S18}
\expandafter\def\csname supplab@sec:clicpro\endcsname{3.8}
\expandafter\def\csname supplab@sec:clicpro@cref\endcsname{[subsection][8][3]3.8}
\expandafter\def\csname supplab@fig:classicfront\endcsname{S3}
\expandafter\def\csname supplab@fig:classicfront@cref\endcsname{[figure][3][]S3}
\expandafter\def\csname supplab@tab:xdevcpu\endcsname{S19}
\expandafter\def\csname supplab@tab:xdevcpu@cref\endcsname{[table][19][]S19}
\expandafter\def\csname supplab@tab:xdevgpu\endcsname{S20}
\expandafter\def\csname supplab@tab:xdevgpu@cref\endcsname{[table][20][]S20}
\expandafter\def\csname supplab@sec:abl\endcsname{4}
\expandafter\def\csname supplab@sec:abl@cref\endcsname{[section][4][]4}
\expandafter\def\csname supplab@fig:perimg\endcsname{S4}
\expandafter\def\csname supplab@fig:perimg@cref\endcsname{[figure][4][]S4}
\expandafter\def\csname supplab@tab:classic\endcsname{S21}
\expandafter\def\csname supplab@tab:classic@cref\endcsname{[table][21][]S21}
\expandafter\def\csname supplab@fig:abl_full\endcsname{S5}
\expandafter\def\csname supplab@fig:abl_full@cref\endcsname{[figure][5][]S5}
\expandafter\def\csname supplab@sec:supp_decomp\endcsname{4.2}
\expandafter\def\csname supplab@sec:supp_decomp@cref\endcsname{[subsection][2][4]4.2}
\expandafter\def\csname supplab@tab:deccost\endcsname{S22}
\expandafter\def\csname supplab@tab:deccost@cref\endcsname{[table][22][]S22}
\expandafter\def\csname supplab@tab:gpumem\endcsname{S23}
\expandafter\def\csname supplab@tab:gpumem@cref\endcsname{[table][23][]S23}
\expandafter\def\csname supplab@tab:bam\endcsname{S24}
\expandafter\def\csname supplab@tab:bam@cref\endcsname{[table][24][]S24}
\expandafter\def\csname supplab@tab:clicpro\endcsname{S25}
\expandafter\def\csname supplab@tab:clicpro@cref\endcsname{[table][25][]S25}
\expandafter\def\csname supplab@fig:loo\endcsname{S6}
\expandafter\def\csname supplab@fig:loo@cref\endcsname{[figure][6][]S6}
\expandafter\def\csname supplab@tab:abl\endcsname{S26}
\expandafter\def\csname supplab@tab:abl@cref\endcsname{[table][26][]S26}
\expandafter\def\csname supplab@sec:v2\endcsname{4.3}
\expandafter\def\csname supplab@sec:v2@cref\endcsname{[subsection][3][4]4.3}
\expandafter\def\csname supplab@tab:ledger\endcsname{S27}
\expandafter\def\csname supplab@tab:ledger@cref\endcsname{[table][27][]S27}
\expandafter\def\csname supplab@tab:loo\endcsname{S28}
\expandafter\def\csname supplab@tab:loo@cref\endcsname{[table][28][]S28}
\expandafter\def\csname supplab@sec:supp_data\endcsname{4.4}
\expandafter\def\csname supplab@sec:supp_data@cref\endcsname{[subsection][4][4]4.4}
\expandafter\def\csname supplab@fig:affine\endcsname{S7}
\expandafter\def\csname supplab@fig:affine@cref\endcsname{[figure][7][]S7}
\expandafter\def\csname supplab@tab:decomp\endcsname{S29}
\expandafter\def\csname supplab@tab:decomp@cref\endcsname{[table][29][]S29}
\expandafter\def\csname supplab@tab:featin\endcsname{S30}
\expandafter\def\csname supplab@tab:featin@cref\endcsname{[table][30][]S30}
\expandafter\def\csname supplab@tab:intcost\endcsname{S31}
\expandafter\def\csname supplab@tab:intcost@cref\endcsname{[table][31][]S31}
\expandafter\def\csname supplab@fig:data\endcsname{S8}
\expandafter\def\csname supplab@fig:data@cref\endcsname{[figure][8][]S8}
\expandafter\def\csname supplab@tab:v2\endcsname{S32}
\expandafter\def\csname supplab@tab:v2@cref\endcsname{[table][32][]S32}
\expandafter\def\csname supplab@tab:data\endcsname{S33}
\expandafter\def\csname supplab@tab:data@cref\endcsname{[table][33][]S33}
\expandafter\def\csname supplab@sec:persist\endcsname{4.6}
\expandafter\def\csname supplab@sec:persist@cref\endcsname{[subsection][6][4]4.6}
\expandafter\def\csname supplab@tab:hyper\endcsname{S34}
\expandafter\def\csname supplab@tab:hyper@cref\endcsname{[table][34][]S34}
\expandafter\def\csname supplab@fig:bufschema\endcsname{S9}
\expandafter\def\csname supplab@fig:bufschema@cref\endcsname{[figure][9][]S9}
\expandafter\def\csname supplab@tab:c1\endcsname{S35}
\expandafter\def\csname supplab@tab:c1@cref\endcsname{[table][35][]S35}
\expandafter\def\csname supplab@sec:bufvar\endcsname{4.6}
\expandafter\def\csname supplab@sec:bufvar@cref\endcsname{[subsection][6][4]4.6}
\expandafter\def\csname supplab@fig:zerok\endcsname{S10}
\expandafter\def\csname supplab@fig:zerok@cref\endcsname{[figure][10][]S10}
\expandafter\def\csname supplab@sec:needs\endcsname{4.7}
\expandafter\def\csname supplab@sec:needs@cref\endcsname{[subsection][7][4]4.7}
\expandafter\def\csname supplab@sec:adapttests\endcsname{4.8}
\expandafter\def\csname supplab@sec:adapttests@cref\endcsname{[subsection][8][4]4.8}
\expandafter\def\csname supplab@sec:mechmaps\endcsname{4.9}
\expandafter\def\csname supplab@sec:mechmaps@cref\endcsname{[subsection][9][4]4.9}
\expandafter\def\csname supplab@tab:persist\endcsname{S36}
\expandafter\def\csname supplab@tab:persist@cref\endcsname{[table][36][]S36}
\expandafter\def\csname supplab@sec:theory\endcsname{5}
\expandafter\def\csname supplab@sec:theory@cref\endcsname{[section][5][]5}
\expandafter\def\csname supplab@tab:obs\endcsname{S37}
\expandafter\def\csname supplab@tab:obs@cref\endcsname{[table][37][]S37}
\expandafter\def\csname supplab@fig:mechgrid\endcsname{S11}
\expandafter\def\csname supplab@fig:mechgrid@cref\endcsname{[figure][11][]S11}
\expandafter\def\csname supplab@fig:texdec\endcsname{S12}
\expandafter\def\csname supplab@fig:texdec@cref\endcsname{[figure][12][]S12}
\expandafter\def\csname supplab@eq:th_h1\endcsname{S1}
\expandafter\def\csname supplab@eq:th_h1@cref\endcsname{[equation][1][]S1}
\expandafter\def\csname supplab@eq:th_h2\endcsname{S2}
\expandafter\def\csname supplab@eq:th_h2@cref\endcsname{[equation][2][]S2}
\expandafter\def\csname supplab@eq:th_q\endcsname{S3}
\expandafter\def\csname supplab@eq:th_q@cref\endcsname{[equation][3][]S3}
\expandafter\def\csname supplab@eq:th_z\endcsname{S4}
\expandafter\def\csname supplab@eq:th_z@cref\endcsname{[equation][4][]S4}
\expandafter\def\csname supplab@sec:tied\endcsname{5.3}
\expandafter\def\csname supplab@sec:tied@cref\endcsname{[subsection][3][5]5.3}
\expandafter\def\csname supplab@sec:stale\endcsname{5.4}
\expandafter\def\csname supplab@sec:stale@cref\endcsname{[subsection][4][5]5.4}
\expandafter\def\csname supplab@tab:depth\endcsname{S38}
\expandafter\def\csname supplab@tab:depth@cref\endcsname{[table][38][]S38}
\expandafter\def\csname supplab@tab:depthad\endcsname{S39}
\expandafter\def\csname supplab@tab:depthad@cref\endcsname{[table][39][]S39}
\expandafter\def\csname supplab@eq:th_adapt\endcsname{S5}
\expandafter\def\csname supplab@eq:th_adapt@cref\endcsname{[equation][5][]S5}
\expandafter\def\csname supplab@sec:vstheory\endcsname{6}
\expandafter\def\csname supplab@sec:vstheory@cref\endcsname{[section][6][]6}
\expandafter\def\csname supplab@sec:unimodal\endcsname{6.1}
\expandafter\def\csname supplab@sec:unimodal@cref\endcsname{[subsection][1][6]6.1}
\expandafter\def\csname supplab@eq:th_unimodal\endcsname{S6}
\expandafter\def\csname supplab@eq:th_unimodal@cref\endcsname{[equation][6][]S6}
\expandafter\def\csname supplab@eq:th_mixture\endcsname{S7}
\expandafter\def\csname supplab@eq:th_mixture@cref\endcsname{[equation][7][]S7}
\expandafter\def\csname supplab@sec:lattice\endcsname{6.2}
\expandafter\def\csname supplab@sec:lattice@cref\endcsname{[subsection][2][6]6.2}
\expandafter\def\csname supplab@sec:vs_res\endcsname{6.3}
\expandafter\def\csname supplab@sec:vs_res@cref\endcsname{[subsection][3][6]6.3}
\expandafter\def\csname supplab@fig:lattice\endcsname{S13}
\expandafter\def\csname supplab@fig:lattice@cref\endcsname{[figure][13][]S13}
\expandafter\def\csname supplab@tab:vsmech\endcsname{S40}
\expandafter\def\csname supplab@tab:vsmech@cref\endcsname{[table][40][]S40}
\expandafter\def\csname supplab@tab:vscfg\endcsname{S41}
\expandafter\def\csname supplab@tab:vscfg@cref\endcsname{[table][41][]S41}
\expandafter\def\csname supplab@sec:vsdecomp\endcsname{6.4}
\expandafter\def\csname supplab@sec:vsdecomp@cref\endcsname{[subsection][4][6]6.4}
\expandafter\def\csname supplab@eq:th_vsg\endcsname{S8}
\expandafter\def\csname supplab@eq:th_vsg@cref\endcsname{[equation][8][]S8}
\expandafter\def\csname supplab@eq:th_vsdecomp\endcsname{S9}
\expandafter\def\csname supplab@eq:th_vsdecomp@cref\endcsname{[equation][9][]S9}
\expandafter\def\csname supplab@tab:vsdecomp\endcsname{S42}
\expandafter\def\csname supplab@tab:vsdecomp@cref\endcsname{[table][42][]S42}
\expandafter\def\csname supplab@sec:mix\endcsname{6.5}
\expandafter\def\csname supplab@sec:mix@cref\endcsname{[subsection][5][6]6.5}
\expandafter\def\csname supplab@tab:mix\endcsname{S43}
\expandafter\def\csname supplab@tab:mix@cref\endcsname{[table][43][]S43}

\newcommand{\sref}[1]{\ifcsname supplab@#1\endcsname\csname supplab@#1\endcsname\else\textbf{??}\fi}

\newcommand{\gumus}{G\"{u}m\"{u}\c{s} and Kam{\i}\c{s}l{\i}}
\newcommand{\gumusShort}{G\"{u}m\"{u}\c{s} et al.}
\providecommand{\cref}[1]{\ref{#1}}

\title{PerLIC: Learned Lossless Compression with Small Integer CPU Decoders}
\author{Daxin Li \enskip Kai Wang \enskip Yuanchao Bai\thanks{Corresponding author.} \enskip Yujia Chen \enskip Wenbo Zhao \enskip Junjun Jiang \enskip Xianming Liu\\
School of Computer Science and Technology, Harbin Institute of Technology, Harbin, China\\
{\tt\small hahalidaxin@stu.hit.edu.cn, yuanchao.bai@hit.edu.cn}
}
\date{}

\begin{document}

\maketitle
\begin{abstract}
Classical lossless codecs such as JPEG XL decode quickly and exactly on a CPU, which every device has.
The best learned codecs reach lower bitrates but run floating-point networks on a GPU; a probability mismatch with the encoder can then break decoding, and a network small enough to decode in integers at classical speed has little capacity.
We present PerLIC, a learned codec that decodes like a classical one: an eight-bit network with \finSpar k to \finLpar k parameters reads classical features and decodes entirely in integers on a CPU, while a GPU encoder fits and transmits a low-rank adapter for each image.
A recurrent per-pixel state reuses the computation of earlier pixels; under exact decoding, in our runs its short unrolls pay off only from stored states, not from zeros, and a persistent state table supplies these cheaply during fitting.
Decoder-side value sets recover the probability that a predicted distribution with few peaks spreads over the gaps between the values that a channel takes given the earlier channels, with only a per-image switch in the header.
Our smallest model, PerLIC-S, has an \finSvsJxl\% lower bitrate than JPEG XL on \nSeven\ images of seven test sets and decodes about as fast on eight AVX-512 CPU cores, with \finSmac k multiply-adds per pixel and \finSencS\ s of encoding per image.
Our largest model, PerLIC-L, matches the strongest other learned codec that we evaluated with \finRatioCallicL$\times$ fewer multiply-adds.
Every stream decodes bit-exactly, and tested subsets also do so on \xdevCpuN\ CPU models, including Apple ARM, and on \xdevGpuN\ GPU models.
\end{abstract}
\begin{figure}[t]
\centering
\includegraphics[width=\linewidth]{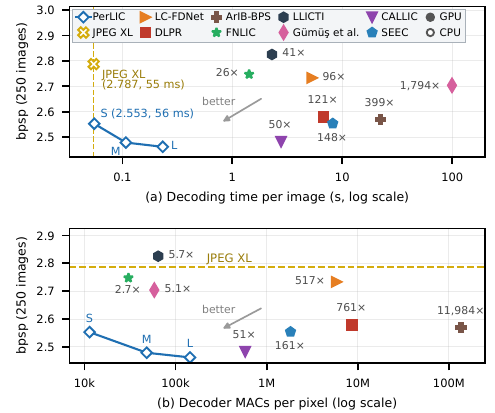}
\caption{Bitrate (bits per sub-pixel, bpsp; \nSeven\ images of seven test sets) against (a) decoding time per image and (b) decoder multiply-accumulate operations (MACs) per pixel (Table~\ref{tab:ext}). Filled: GPU decoding; hollow: 8 CPU cores; dashed: JPEG XL~\cite{alakuijala2019jpeg}. Labels: decoding time (a) or MACs (b) as multiples of PerLIC-S (time with value sets); times of released implementations. Not shown: L3C~\cite{mentzer2019practical}, RC~\cite{mentzer2020learning} (above 3.0 bpsp), OmniZip~\cite{zhao2026omnizip} (code length only), and paq8px~\cite{paq8px} (Supp.~Table~\sref{tab:classic}).}
\label{fig:macs}
\end{figure}

\section{Introduction}
\label{sec:intro}

Lossless compression serves images that must be restored pixel for pixel, such as histology slides, aerial imagery, scanned maps, and photographs.
An image is often encoded once and decoded many times, on whichever device opens it, so a codec is deployed through its decoder, which must be fast and give the same result on every device; every device has a CPU, whereas accelerators differ in availability and arithmetic~\cite{parnamaa2026mlvc}.
Classical codecs such as JPEG XL~\cite{alakuijala2019jpeg} decode quickly and exactly on a CPU.
Learned codecs reach lower bitrates~\cite{salimans2016pixelcnn++,hoogeboom2019integer,callic}, and learned lossy codecs are now being built for such deployment~\cite{parnamaa2026mlvc,jia2026pulse}.
Learned lossless codecs have not followed: the strongest that we evaluate run floating-point networks of \macCallic k to \macAribM M multiply-accumulate operations (MACs) per pixel on a GPU (Figure~\ref{fig:macs}b), and to our knowledge none decodes in integers on a CPU at about the speed of JPEG XL.

Three difficulties stand in the way.
First, the network sets the probability of every pixel, and one differing entry of a probability table can desynchronize the entropy decoder, yet floating-point results depend on the order of additions, which differs between kernels, batch shapes, and devices~\cite{balle2019integer,arunruangsirilert2026streamable,parnamaa2026mlvc}.
Integer arithmetic with fixed rounding removes this risk, but a network that keeps pace with JPEG XL on a CPU can afford only a small fraction of these MACs and has little capacity.
Fitting a model to each image restores capacity, as CALLIC~\cite{callic} and FNLIC~\cite{fnlic} show with floating-point decoders, but FNLIC takes about \fnlicEncH\ hours per image (Table~\ref{tab:ext}), and parameters fitted in floating point must still hold under integer decoding.
Second, such a network affords only a few layers per pixel.
A recurrent state would let every pixel reuse the layers that earlier pixels evaluated, but each state then depends, through its neighbors, on pixels far back: a gradient step that follows this recursion exactly is slow because the recursion is sequential, and a step that cuts it short must start from states that it did not compute, where zeros give the network inputs unlike those of decoding.
Third, a parametric distribution with one or a few peaks is cheap to decode, whereas in some kinds of images a channel takes only a few values given the channels decoded before it; the distribution then puts probability on the gaps between these values, and sending the values instead costs side information.

\begin{figure}[t]
\centering
\includegraphics[width=\linewidth]{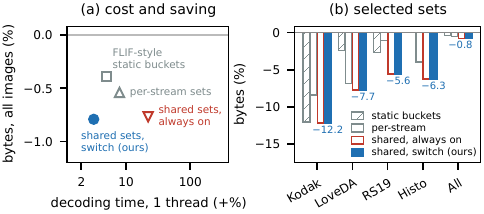}
\caption{Value sets on PerLIC-S (1600 steps, three seeds, actual bitstreams). (a) Change in bytes over all \nSeven\ images against one-thread decoding time, both relative to no sets (24 timing images, separate session). (b) The four configurations on selected test sets and all images (per-stream: one set per rANS stream). Switch: sets only where they save at least 1\%. Static buckets: whole-image sets in the header, after FLIF~\cite{sneyers2016flif}. Values: Supp.~Table~\sref{tab:vs}.}
\label{fig:vs}
\end{figure}

To address these difficulties, we propose PerLIC, a learned lossless codec that decodes like a classical one.
Its decoder is a small eight-bit integer (int8) network that reads integer features of the causal neighborhood (the pixels already decoded), carries a recurrent state from pixel to pixel, and codes with fixed integer tables, entirely on a CPU.
A GPU encoder restores capacity for each image with adapter fitting, which fits a low-rank adapter~\cite{hu2021lora}, and head refit, which refits the output layers on the exact integer activations of decoding, and it sends the fitted parameters.
Two modules address the second and third difficulties.
The \emph{persistent state table} makes fitting through the recurrent state affordable.
It holds the latest state computed for every pixel, so a fitting step recomputes only the states next to its sampled pixels and takes the states beyond them from the table, rather than from zeros or from the sequential recursion.
The table is not sent, and decoding runs the exact recursion without it.
\emph{Decoder-side value sets} target the gaps between channel values without sending the values.
The decoder collects the values that a channel has taken under each key formed from the earlier channels and codes whether the next value is among them with an adaptive escape, moving probability from the gaps to the values seen so far.
Since the sets are built from decoded pixels, the file carries only a per-image switch, which the encoder turns on where the sets save bits.
Our contributions are:
\begin{itemize}
\item A learned lossless codec that decodes like a classical one: an int8 decoder of \finSpar k to \finLpar k parameters with classical integer features and affine color conditioning, and a GPU encoder for adapter fitting and head refit. PerLIC-S is \finSvsJxl\% smaller than JPEG XL on seven test sets, at about its decoding time on eight AVX-512 cores and \finSencS\ s of encoding per image, and PerLIC-L matches CALLIC with \finRatioCallicL$\times$ fewer MACs (Table~\ref{tab:ext}).
\item A recurrent state under exact decoding: at about equal MACs it saves more than explicit depth. In our runs, its short unrolls pay off not from zeros but only from stored states, which a persistent state table supplies in adapter fitting; at the compute of the table, truncated backpropagation through time (TBPTT)~\cite{williams1990bptt} along the exact recursion leaves the files \tabTbpeVsNonrec\% larger than with an adapted non-recurrent state (Section~\ref{sec:analysis}).
\item Decoder-side value sets for parametric heads, with no side information beyond a per-image switch in the header. A head with few peaks, provably a log-concave one or a mixture of too few log-concave components, wastes probability on the gaps between the few values that a channel takes given the earlier channels (Supp.~Sec.~\sref{sec:unimodal}). The saving of the sets comes almost entirely from the head's redundancy on set membership (Section~\ref{sec:analysis}). They save \vsAllGain\% over all images, \vsKodakGain\% on Kodak~\cite{kodak}, and \hiVsGain\% on Histo~\cite{fnlic} (Figure~\ref{fig:vs}).
\end{itemize}

\section{Related Work}
\label{sec:related}

\paragraph{Classical and learned lossless codecs.}
LOCO-I and CALIC predict each pixel from a causal neighborhood and model the residual with context-dependent statistics~\cite{weinberger2000loco,wu1997context}, and TMW, MRP, FLIF, and JPEG XL add stronger predictors and context modeling~\cite{meyer1997tmw,matsuda2005mrp,sneyers2016flif,alakuijala2019jpeg}.
FELICS codes whether a pixel lies in an interval set by two neighbors~\cite{howard1993felics}, PAQ mixes the bit predictions of many context models with small online networks~\cite{mahoney2013dce}, and histogram packing helps on images whose intensities leave gaps~\cite{ferreira2002histogram}. FLIF limits each color to a bucket of values that the encoder sends~\cite{sneyers2016flif}; PPM codes a symbol among those already seen in its context, with an escape~\cite{cleary1984ppm}; the binary mode of CALIC codes a pixel as one of two neighborhood values or an escape~\cite{wu1997context}. Our value sets (Section~\ref{sec:vs}) apply this escape to the values of a channel under each key of the earlier channels (blue under red and green in our codec), collected by the decoder, with a per-image switch in the header.
Autoregressive networks code pixel by pixel~\cite{van2016pixel,salimans2016pixelcnn++}; flows~\cite{hoogeboom2019integer,ho2019compression,van2020idf++,zhang2021ivpf,zhang2021iflow} and bits-back latent models~\cite{townsend2019practical,kingma2019bit,townsend2019hilloc} code many pixels at once, language models code pixels as tokens~\cite{deletang2024language,chen2025p2llm,du2025llm,yu2026wavellm,xu2026shic}, and multi-scale, frequency-based, and lossy-plus-residual designs trade rate for speed~\cite{mentzer2019practical,mentzer2020learning,cao2020lossless,rhee2022lc,kang2022pilc,bai2024deep,Zhang2024ArIBBPS}.
Lighter learned codecs use tens to hundreds of thousands of parameters in floating point~\cite{gumus2024learned,kamisli2023interpolation}, trade rate against MACs~\cite{lopes2024ratecomplexity}, or run integer-only flows on a GPU~\cite{wang2022iodf}.
Our work keeps the autoregressive structure and shrinks the decoder to an integer-only multilayer perceptron that runs on a CPU.

\paragraph{Low-complexity and deterministic decoding.}
Overfitted codecs fit a small decoder to each image~\cite{ladune2023coolchic,leguay2023lowcomplexity,kim2024c3,blard2024overfitted}, down to a few hundred multiplications per pixel.
Floating-point differences between devices can break the decoding of learned entropy models~\cite{balle2019integer,he2022ptq}.
Integer networks~\cite{balle2019integer}, quantized entropy models~\cite{sun2020fixedpoint,koyuncu2022device,he2022ptq}, and quantized decoders~\cite{koyuncu2024quantized} make lossy codecs deterministic; one keeps its floating-point rescaling on the CPU~\cite{tatwawadi2026pico}, and a recent lossy codec decodes on one CPU thread~\cite{jia2026pulse}.
Our decoder is integer-only by design, including the state and the entropy tables, and it codes with interleaved streams of range asymmetric numeral systems (rANS)~\cite{duda2013ans,giesen2014interleaved}.

\paragraph{Instance adaptation.}
Content-adaptive optimization refines the latents or the model for each image~\cite{campos2019content,yang2020improving,vanrozendaal2021overfitting}, and adapters or low-rank updates sent as side information adapt a decoder cheaply~\cite{hu2021lora,tsubota2023universal,shen2023dec,lv2023dynamic,chen2025tta,callic}. We fit such an adapter to an integer decoder through its recurrent state, which the persistent state table makes affordable.

\paragraph{Recurrent state and stored states.}
Recurrent pixel models such as PixelRNN, multi-dimensional RNNs, and spatial LSTMs pass a hidden state along the scan or the diagonals~\cite{van2016pixel,graves2007multi,theis2015generative}, and are trained by backpropagation through time, often truncated~\cite{williams1990bptt}.
Persistent states are closest to the historical embeddings of graph networks~\cite{chen2018vrgcn,fey2021gnnautoscale}, persistent Markov chains~\cite{tieleman2008pcd}, and the stored recurrent states of replay buffers; R2D2 found that unrolls from zero states train worse than unrolls from stored states~\cite{kapturowski2019r2d2}. Exact decoding exposes this mismatch in a causal image model, where zero-state unrolls of up to \bufKmax\ steps and a running-mean state (Supp.~Sec.~\sref{sec:persist}) do not remove it without adapter fitting; we use stored states to adapt an integer recurrent codec, at a fraction of the cost of exact TBPTT.

\section{Method}
\label{sec:method}

\begin{figure*}[t]
\centering
\includegraphics[width=\linewidth]{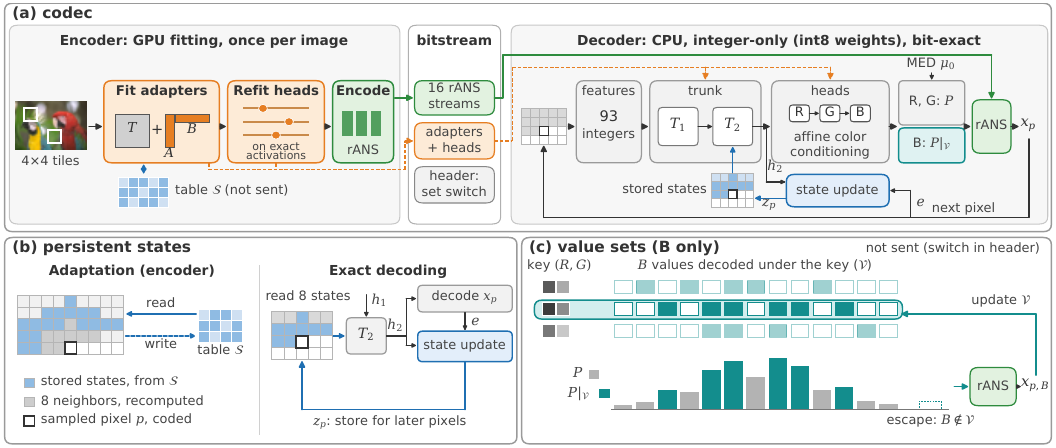}
\caption{Our codec (blue: states; orange: adapted and sent; teal: value sets; green: entropy coding). (a) The GPU encoder fits low-rank adapters, refits the heads, and sends both; the integer decoder predicts an offset $o_0$ to the MED prediction $\mu_0$ and a log scale $o_1$ from the causal neighborhood and eight neighbor states, and updates the state $z_p$ after decoding. (b) Adaptation recomputes the eight neighbor states from states in an unsent table $\mathcal{S}$; decoding runs the exact recursion. (c) $P|_{\mathcal{V}}$: $P$ renormalized over B values $\mathcal{V}$ decoded under the same R and G, with an escape.}
\label{fig:overview}
\end{figure*}

\subsection{Integer-only predictor with a recurrent state}
\label{sec:model}
\paragraph{Decoding order and features.}
Figure~\ref{fig:overview} gives an overview. Pixels are decoded in wavefronts $\tau=x+4y$: every row lags the row above by four pixels because the features read the row above up to three pixels to the right, so all pixels with the same $\tau$ are independent.
For the three channels $c$ of pixel $p$, the median edge detector (MED) of LOCO-I~\cite{weinberger2000loco} gives a prediction $\mu_{0,c}$ in Q6 fixed point (six fractional bits) from the causal neighbors.
We build 93 integer features $t_p$ per pixel from the causal neighborhood. They are pixel differences in a causal window of $p$, the classical residuals at nearby decoded positions, and 33 values from a wider window above $p$. They also include further classical predictions, namely the gradient-adjusted predictor of CALIC, a least-squares predictor, and the north and west pixels. Supp.~Table~\sref{tab:feat} lists all 93.

\paragraph{Integer layers and recurrent state.}
A trunk layer $T$ (any layer before the heads) has int8 weights $W$, an int32 bias $b$, and an integer multiplier $\kappa$ with 24 fractional bits per output row and computes
\begin{equation}
\mathrm{clip}(Tx)=\Big[\big\lfloor\big((Wx+b)\,\kappa+2^{23}\big)/2^{24}\big\rfloor\Big]_0^{255}
\label{eq:intlayer}
\end{equation}
in integer arithmetic only, with clamping to $[0,255]$ (Supp.~Sec.~\sref{sec:net}).
The first layer is $h_1=\mathrm{clip}(T_1 t_p)\in\mathbb{Z}_{0..255}^{H}$, where the hidden width $H$ is 32 to 256.
The state of pixel $p$ is $z_p\in\mathbb{Z}_{0..255}^{16}$.
The second layer reads $h_1$ and the states of the eight causal neighbors $N_1,\ldots,N_8$ of $p$ (W, N, NW, NE, WW, NN, NWW, and NEE),
\begin{equation}
h_2=\mathrm{clip}\big(T_2\,[\,h_1;\,z_{N_1};\ldots;z_{N_8}\,]\big).
\label{eq:h2}
\end{equation}
The state of $p$ is computed with weights $U$ and $U_e$ after all three channels of $p$ are decoded,
\begin{equation}
z_p=\mathrm{clip}\big(U\,h_2+U_e\,[\,e;\,|e|\,]\big),\qquad e_c=64\,x_c-\mu_{0,c}.
\label{eq:state}
\end{equation}
Here $e$, the classical residual of the decoded pixel in Q6 and the only input of the state that depends on $x_p$, is the \emph{residual observation}. Like the context bias and activity statistics of LOCO-I~\cite{weinberger2000loco}, the state accumulates prediction errors, but over the causal neighborhood, and adaptation calibrates their readout. Reading $h_2$ makes the state recurrent; a \emph{non-recurrent state} reads $h_1$ in~\eqref{eq:state}, and~\eqref{eq:h2} still reads neighbor states. Unrolled, $z_p$ is the output of a weight-tied causal network of $d(p)+1$ state-update blocks, where $d(p)$ is the length of the longest chain of state neighbors that ends in $p$: each pixel evaluates one block and shares it with up to eight later pixels (Supp.~Sec.~\sref{sec:tied}).

\paragraph{Heads and color conditioning.}
Small output heads, one per channel (linear for red, 32 hidden units for green and blue), map $h_2$ to a mean offset $o_{0,c}$ and a log scale $o_{1,c}$. The colors are coded in the order R, G, B, and each follows a discretized logistic distribution,
\begin{equation}
P(x_c{=}v)=\sigma\Big(\frac{v+\frac12-m_c}{s_c}\Big)-\sigma\Big(\frac{v-\frac12-m_c}{s_c}\Big),
\label{eq:logistic}
\end{equation}
where $\sigma$ is the logistic function and the tails are merged into $v=0$ and $v=255$.
Green and blue depend on two errors of each earlier color $j<c$ of the same pixel, the head error $\eta_{j,1}=x_j-m_j$ and the MED residual $\eta_{j,2}=e_j/64$, through \emph{affine color conditioning}: from $h_2$ alone, their heads also output coefficients $\alpha_{k,c,j,a}$, and
\begin{equation}
\tilde o_{k,c}=o_{k,c}+\sum_{j<c}\sum_{a=1}^{2}\alpha_{k,c,j,a}\,\phi_k(\eta_{j,a}),\quad k\in\{0,1\},
\label{eq:affine}
\end{equation}
with $\phi_0(\eta)=\eta$, $\phi_1(\eta)=|\eta|$ (fixed scales folded into $\alpha$), $m_c=\mu_{0,c}/64+\tilde o_{0,c}/16$, and $s_c=\exp(\tilde o_{1,c})$ in~\eqref{eq:logistic}.
The trunk and heads thus run before red is decoded, and only these few MACs wait for a decoded color. PixelCNN++~\cite{salimans2016pixelcnn++} shifts the means of green and blue by multiples of the decoded values; we use prediction errors instead, also adapt the scale, and keep the coefficients on an integer grid.
Training and adaptation minimize the code length $L=-\sum_p\sum_c\log_2P(x_{p,c})$.

\paragraph{Entropy coding and model sizes.}
The decoder reads $P$ from a fixed integer table whose cumulative frequencies total $2^{20}$ (Supp.~Sec.~\sref{sec:p20}) and codes each symbol under $P$ with rANS~\cite{duda2013ans} using 16 interleaved streams, so the decoder can use several threads.
PerLIC-S has $H=32$, \finSpar k parameters, and \finSmac k MACs per pixel. PerLIC-M has $H=96$ and a third fully connected layer of 96 units, read by the state and the heads (\finMmac k MACs), and PerLIC-L has $H=256$ (\finLmac k); we write S, M, and L for short. The counts cover the network; the features add 33 multiplications, which the decoding times include (Supp.~Table~\sref{tab:net}). Our timed decoder (fast path) uses AVX-512 VNNI dot products and decodes two rows at once without changing the output (Supp.~Sec.~\sref{sec:decoder}).

\subsection{Per-image adaptation with a persistent state table}
\label{sec:adapt}
Per-image adaptation has two stages (Figure~\ref{fig:overview}a): adapter fitting through the recurrent state with a persistent state table, then head refit on the exact integer recursion.

\paragraph{Adapter fitting with persistent states.}
Adapter fitting minimizes the code length $L(x;\theta+\Delta)$ of the image over $\Delta$: low-rank increments $T_i+A_iB_i$ ($i=1,2$, and 3 in M; the rank grows with the image size, Supp.~Sec.~\sref{sec:adapt}) and increments of the output-layer weights and biases. It runs $K_A$ Adam~\cite{kingma2015adam} steps ($K_A$ is 400, 800, or 1600 below, followed by head refit unless stated) with a linearly decaying learning rate, and every step samples tiles of $4\times4$ pixels.
The state of a pixel depends on the states of its neighbors, which depend on theirs, so a step on sampled pixels can unroll the recursion only a few hops (state offsets) back.
With weights $\theta$ and adapter $\Delta$, we write $z_q=f_{\theta+\Delta}(q,z_{N(q)})$ for the state update~\eqref{eq:state}, where $z_{N(q)}$ are the states of the eight causal neighbors of $q$.
For a sampled pixel $p$, a step computes
\begin{equation}
z^{(0)}_r=\mathcal{S}[r],\qquad z^{(k)}_q=f_{\theta+\Delta}\big(q,z^{(k-1)}_{N(q)}\big),\quad k=1,\ldots,K,
\label{eq:unroll}
\end{equation}
on the pixels that are $K-k+1$ hops from $p$, and codes $p$ with the neighbor states $z^{(K)}_{N(p)}$.
The states $z^{(0)}$, $K+1$ hops away, must come from elsewhere; zeros would give the network inputs unlike those in decoding (Section~\ref{sec:analysis}; Supp.~Table~\sref{tab:persist}).
The \emph{persistent state table} $\mathcal{S}$ of an image instead holds the latest state of every pixel, stale once the weights change. It is initialized by the exact recursion with the unadapted weights, and after each step $\mathcal{S}[q]\gets\mathrm{sg}(z_q)$ stores the new states, where sg stops the gradient.
Adapter fitting uses $K=1$ (training: $K=2$), since the table supplies the deeper states, so the loss passes through the eight neighbor states, recomputed with the current weights and adapter (Supp.~Alg.~\sref{alg:enc}).
\paragraph{Head refit.}
The decoder runs the exact integer recursion, but adapter fitting sees approximate activations (stale table states, floating point, unrounded factors $A_i,B_i$), so heads fitted to them are miscalibrated for decoding. With the factors quantized and merged into the int8 weights, we refit the output layers of the heads for 300 steps on the exact integer activations, which stay fixed because the heads do not feed the states; the refit thus costs one integer pass and a fit of the output layers. \emph{Head refit only} skips adapter fitting. The factors and output layers are sent as side information (Supp.~Table~\sref{tab:sideinfo}). Our encoder runs each step as a compiled CUDA graph\label{sec:fast}, which changes the result only by floating-point rounding (Supp.~Sec.~\sref{sec:encab}).

\subsection{Decoder-side value sets}
\label{sec:vs}
A head with few peaks, such as our discretized logistic $P$, puts mass on the gaps between the few values that a channel may take given the earlier channels (Supp.~Sec.~\sref{sec:unimodal}).
We apply value sets to blue only (Supp.~Sec.~\sref{sec:vs_res}).
The decoder keeps, for every key, the set $\mathcal{V}$ of the blue values decoded under that key, counting only rows at least 16 above (always finished) and the current row so far, so that $\mathcal{V}$ does not depend on the thread schedule. The key is the pixel's red and green values, exact (exact key) or with their low bits dropped (coarse key), which pools similar colors when an image has too few pixels per exact key.
For a key with a small non-empty set and an escape probability $\varepsilon$, blue is coded with
\begin{equation}
P|_{\mathcal{V}}(v)=\begin{cases}(1-\varepsilon)\,P(v)/P(\mathcal{V}), & v\in\mathcal{V},\\ \varepsilon\,P(v)/\big(1-P(\mathcal{V})\big), & v\notin\mathcal{V},\end{cases}
\label{eq:pv}
\end{equation}
as a binary in-or-out flag followed by the value under $P$ renormalized inside or outside $\mathcal{V}$ ($P(\mathcal{V})=\sum_{v\in\mathcal{V}}P(v)$); otherwise blue is coded with $P$.
A value in $\mathcal{V}$ thus costs $\log_2\frac{1-\varepsilon}{P(\mathcal{V})}$ bits less than under $P$, and a new value costs $\log_2\frac{1-P(\mathcal{V})}{\varepsilon}$ bits more.
In each context, given by the size of $\mathcal{V}$ and the head's escape mass $1-P(\mathcal{V})$, $\varepsilon$ starts at that mass and follows the outcomes as an integer moving average.
The 16 streams share the sets, each with its own escape contexts. The encoder writes the file with exact keys, with coarse keys, and without sets, and keeps the smaller file with sets only if it saves at least 1\%; the header records the choice (integer rule and constants: Supp.~Sec.~\sref{sec:vs_supp}).

\section{Experiments}
\label{sec:exp}
\begin{table*}[t]
\centering\small
\resizebox{\ifdim\width>\textwidth \textwidth\else\width\fi}{!}{\begin{tabular}{@{}lcccccccccccc@{}}
\toprule
 & \multicolumn{8}{c}{bpsp $\downarrow$} & MACs & & \multicolumn{2}{c}{time (s)} \\
\cmidrule(lr){2-9}\cmidrule(lr){12-13}
Method & Kodak & DIV2K & CLIC & Doc & LoveDA & RS19 & Histo & All & /px (k) & Params & Enc & Dec \\
\midrule
\multicolumn{13}{@{}l}{\textit{Classical: tested on 8 \textbf{CPU} threads (library call)}} \\
JPEG XL, effort 10 & 2.865 & 2.780 & 2.737 & 2.295 & 3.315 & 1.946 & 4.082 & 2.787 & -- & -- & 38.57 & \textbf{0.055} \\
\midrule
\multicolumn{13}{@{}l}{\textit{Learned codecs: tested on the \textbf{GPU} (released code)}} \\
L3C~\cite{mentzer2019practical} & 3.263 & 3.127 & 3.098 & 3.274 & 3.745 & 2.673 & 4.541 & 3.165 & 432 & 5.0M & 2.41 & 2.54 \\
RC~\cite{mentzer2020learning} & 3.376 & 3.069 & 3.067 & 3.468 & 3.715 & 2.195 & 4.502 & 3.115 & 3083$^\S$ & 7.4M & 2.30 & 0.87 \\
LC-FDNet~\cite{rhee2022lc} & 2.938 & 2.705 & 2.658 & 2.721 & 3.337 & 1.949 & 4.123 & 2.733 & 5866$^\S$ & 23.8M & 5.28 & 5.32 \\
LLICTI~\cite{kamisli2023interpolation} & 3.013 & 2.775 & 2.791 & 2.866 & 3.488 & 2.264 & 4.223 & 2.825 & 64 & 197k & \textbf{2.14} & 2.29 \\
G\"{u}m\"{u}\c{s} et al.~\cite{gumus2024learned} & 2.876 & 2.657 & 2.667 & 2.622 & 3.434 & 2.025 & 4.205 & 2.704 & 58 & 64k & 15.44 & 99.59 \\
DLPR~\cite{bai2024deep} & 2.864 & 2.537 & 2.484 & 2.683 & 3.279 & 2.008 & 4.127 & 2.579 & 8628 & 15.6M & 2.99 & 6.73 \\
ArIB-BPS~\cite{Zhang2024ArIBBPS} & 2.781 & 2.539 & 2.477 & 2.617 & 3.234 & 1.910 & 3.945 & 2.569 & 135902 & 146.3M & 24.19 & 22.16 \\
FNLIC~\cite{fnlic} & 2.881 & 2.745 & 2.664 & 2.224 & 3.295 & 2.232 & 3.864 & 2.747 & \underline{30} & 83k & 12331 & 1.42 \\
CALLIC~\cite{callic} & 2.590 & \textbf{2.471} & \textbf{2.388} & 2.126 & 3.165 & 1.783 & 3.867 & 2.482 & 580$^\dagger$ & 600k & 26.30 & 2.78 \\
SEEC~\cite{zheng2026seec} & 2.853 & 2.510 & 2.455 & 2.667 & 3.259 & 1.963 & 4.165 & 2.554 & 1827 & 34.9M & 3.85 & 8.19 \\
OmniZip-L$^\ast$~\cite{zhao2026omnizip} & 2.992 & 2.975 & 2.861 & 3.368 & 3.602 & 2.784 & 4.318 & 3.001 & 281143 & 208.3M & 45.62 & -- \\
\midrule
\multicolumn{13}{@{}l}{\textit{PerLIC (ours): encoder on the \textbf{GPU} + \textbf{CPU}, integer decoder on 8 \textbf{CPU} threads}} \\
PerLIC-S & 2.386 & 2.565 & 2.496 & 2.199 & 2.903 & 1.756 & 3.570 & 2.553 & \textbf{11.3} & \textbf{11.5k} & \underline{2.15} & \underline{0.056} \\
PerLIC-M & \underline{2.302} & 2.497 & 2.418 & \underline{1.938} & \textbf{2.886} & \underline{1.650} & \underline{3.520} & \underline{2.479} & 48.1 & \underline{48k} & 3.24 & 0.108 \\
PerLIC-L & \textbf{2.301} & \underline{2.479} & \underline{2.401} & \textbf{1.907} & \underline{2.892} & \textbf{1.611} & \textbf{3.517} & \textbf{2.462} & 143.7 & 144k & 7.14 & 0.233 \\
\bottomrule
\end{tabular}
}
\caption{Bitrate (bpsp) and time per image (s; released implementations, mean over the 24 timing images on the device of the group row; our encoding time includes per-image fitting on the GPU and value-set coding in CPU threads, after an uncounted warm-up image per set for one-off compilation). All: the \nSeven\ images pooled. Bold/underlined: best/second best; --: not applicable. PerLIC: 1600 adaptation steps, value sets; bitrates over three training seeds, times of seed~0. $^\dagger$ Adapter branches not counted. $^\ast$ Theoretical code length, not ranked; time: network pass only. $^\S$ Network only; the classical codec inside (BPG in RC, JPEG XL in LC-FDNet) is timed but not counted.}
\label{tab:ext}
\end{table*}

\subsection{Setup}
\label{sec:setup}
\paragraph{Data.}
We evaluate on seven test sets with \nSeven\ images in total: Kodak (24 images), the DIV2K and CLIC 2024 validation sets (100 and 30), Doc (24), LoveDA (24), RS19 (24), and Histo (24)~\cite{kodak,Agustsson_2017_CVPR_Workshops,clic}.
Doc, LoveDA, and Histo are the $768\times512$ patch sets of FNLIC~\cite{fnlic} (historical maps~\cite{lin2024rumsey}, aerial photographs~\cite{wang2021loveda}, and histology), and RS19 has 24 crops of $576\times576$ pixels from WHU-RS19~\cite{dai2011rs19} (Supp.~Sec.~\sref{sec:data}).
The bitrate is bits per sub-pixel (bpsp) of the complete file, including the header and the adapter side information, pooled over all pixels of the seven sets (DIV2K and CLIC hold \pixShareDivClic\% of them); model weights shared by all images are not counted for any codec.

\paragraph{Baselines.}
JPEG XL uses effort 10 with default settings (effort 7: Supp.~Table~\sref{tab:classic}).
L3C~\cite{mentzer2019practical}, RC~\cite{mentzer2020learning}, LC-FDNet~\cite{rhee2022lc}, LLICTI~\cite{kamisli2023interpolation}, the codec of \gumus~\cite{gumus2024learned}, DLPR~\cite{bai2024deep}, ArIB-BPS~\cite{Zhang2024ArIBBPS}, FNLIC~\cite{fnlic}, CALLIC~\cite{callic}, and SEEC~\cite{zheng2026seec} are learned codecs with floating-point decoders; LLICTI and \gumusShort\ are low-complexity designs with \parLlictiK k and a nominal \parGumusNominal k parameters (\parGumusStored k stored in its dense masked kernels).
We run every external codec once on the same pixels with its released code and weights (FNLIC's RS19 models fitted with its encoder) and count the actual bytes of their streams, including the per-image adapters of CALLIC; OmniZip~\cite{zhao2026omnizip} releases no working decoder, so we report the theoretical code length of its largest model, which favors it, and leave it unranked (details in Supp.~Sec.~\sref{sec:extruns_supp}). No other codec in Table~\ref{tab:ext} uses value sets.
For the ablation, we train our decoder with a non-recurrent state, without residual observation, and with a learned blend of six classical predictors (blended base) or MED as base, all at the MACs of S.

\paragraph{Training.}
We train on random pixels with a batch of 16384 for 160k float and 60k quantization-aware steps, unrolling \eqref{eq:unroll} for $K=2$ steps without an adapter. The states at the truncation are carried over from earlier training steps; they approximate the exact states without reaching them (Supp.~Sec.~\sref{sec:stale}). Truncated backpropagation along the exact recursion reaches about the same bitrate after adaptation (Supp.~Sec.~\sref{sec:persist}).
The training set has \sWebList\ web photos and \sCamKept\ camera photos from \corpusTotal\ public images~\cite{deng2009imagenet,lin2014coco,clic,Agustsson_2017_CVPR_Workshops}, one $512\times512$ crop each, camera photos area-reduced first (Supp.~Sec.~\sref{sec:supp_data}).

\paragraph{Protocol.}
The integer C decoder (scalar path) decodes every stream, and the decoded image must equal the original exactly.
PerLIC is the mean over three training seeds with one adaptation seed. For S, two further adaptation seeds give a standard deviation of \aseedSdAll\% of the bytes (Doc: \aseedSdDoc\%), against \tseedSdAll\% (Doc: \tseedSdDoc\%) over training seeds.
Some mechanism results use the theoretical code length under a stated table (actual bitstreams of S without adaptation exceed it by \realOverIdeal\% with a table of total $2^{16}$).
Times are means over 24 timing images, eight each from Kodak, DIV2K, and CLIC, on one NVIDIA H20 GPU or eight physical cores of an AMD EPYC (Zen 4) server CPU (AVX-512 with VNNI int8 dot-product instructions), with JPEG XL built from the libjxl source for this CPU without profile-guided optimization, and every timed decoding is compared with the original (Supp.~Sec.~\sref{sec:timing}).
\subsection{Compression performance and speed}
\label{sec:perf}
Table~\ref{tab:ext} shows the bitrate of all codecs on the seven test sets and their speed on the timing images (more comparisons and timings: Supp.~Sec.~\sref{sec:res}).
With 1600 adaptation steps and value sets, PerLIC-S is \finSvsJxl\% smaller than JPEG XL (\finSnoVsVsJxl\% without value sets).
It encodes an image in \finSencS\ s on one H20 GPU, value-set coding on the CPU included; fewer steps trade bitrate for time (Supp.~Table~\sref{tab:main_all}). Without adaptation, our one-thread C encoder takes \cpuEncS\ s per image and is \cpuEncVsJxl\% smaller than JPEG XL (Supp.~Sec.~\sref{sec:adapttests}).

\paragraph{Decoding speed.}
On eight cores, S (profile-guided build; \sNopgoMin--\sNopgoMax\% slower without) takes \finSdecEightMs\ ms against \finJxlDecEightMs\ ms for JPEG XL, faster than every learned GPU codec; JPEG XL gains mainly on Kodak, where every image uses value sets (S: \finSdecEightPlainMs\ ms overall without them; Supp.~Table~\sref{tab:dectime}). One thread: \finSdecOneMs\ vs.\ \finJxlDecOneMs\ ms.

\subsection{Comparison with learned codecs}
\label{sec:extcmp}

S needs \finFnlicOverSmac$\times$ fewer MACs than FNLIC and is smaller on every set (by \finSFnlicMin--\finSFnlicMax\%); it is also smaller than \gumusShort, L3C, RC, LC-FDNet, and LLICTI.
On Kodak, LoveDA, RS19, and Histo, S with value sets is smaller than every external learned codec; without them, it already is on LoveDA and Histo. The sets carry over to CALLIC (\cvsKodak\% smaller on Kodak) and make it smaller than S on Kodak and RS19 (Supp.~Sec.~\sref{sec:vs_res}).
On DIV2K and CLIC (large natural photos), DLPR, ArIB-BPS, SEEC, and CALLIC are smaller than S, and CALLIC also on Doc. Pooled over all images, S is slightly smaller than DLPR and ArIB-BPS, about equal to SEEC, and \finSoverCallic\% larger than CALLIC, which needs \finRatioCallicS$\times$ more MACs.
The larger models close this gap: L is \finLvsCallic\% smaller than CALLIC (about equal without value sets) with \finRatioCallicL$\times$ fewer MACs, and M is about equal to it with \finRatioCallicM$\times$ fewer MACs. CALLIC stays smaller than L on DIV2K and CLIC, but L is smaller than every external learned codec on the other five sets. With value sets in both, L and CALLIC are about tied.
Context-mixing codecs fall on both sides of S: paq8px~\cite{paq8px} is smaller but far slower, GraLIC~\cite{rhatushnyak2011lpcb} larger and slower (Supp.~Table~\sref{tab:classic}).
On eight cores, S decodes \finScpuVsCallicMin--\finScpuVsCallicMax$\times$ faster than CALLIC on its GPU (Kodak, DIV2K, and CLIC; Supp.~Table~\sref{tab:dectime}).

\paragraph{Bit-exactness.} Integer arithmetic and a fixed decoding order make the output independent of the vector width, the device, and the thread schedule (Supp.~Sec.~\sref{sec:determ}). Our timed decoders reproduce the scalar path on all \finRegressN\ image--seed--model cases (eight widths; Supp.~Table~\sref{tab:regress}), and every tested stream decodes to the original on \xdevCpuN\ AMD, Intel, and Apple (arm64) CPU models and, with a CUDA build, on \xdevGpuN\ NVIDIA GPU models (value-set streams of S on \xdevVsGpuN; Supp.~Sec.~\sref{sec:xdev}). Without adaptation, integer decoding of S costs \sIntCostTwenty\% against its float network after float training (Supp.~Table~\sref{tab:intcost}).

\subsection{Analysis}
\label{sec:analysis}
All analyses below use PerLIC-S on the \nSeven\ images, with three training seeds unless stated, and actual bitstreams without value sets; adaptation uses 800 steps unless stated.
\begin{figure}[t]
\centering
\includegraphics[width=\linewidth]{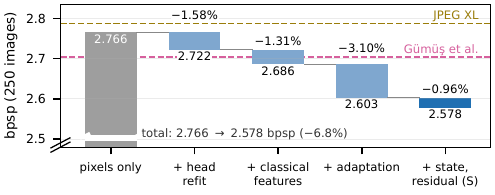}
\caption{Gain ledger of PerLIC-S (\nSeven\ images, three seeds, actual bitstreams), adding the parts in the order of Section~\ref{sec:analysis}, with 800 adaptation steps and no value sets; the shares depend on this order (per set: Supp.~Table~\sref{tab:ledger}; variants: Supp.~Table~\sref{tab:decomp}). Dashed: JPEG XL and \gumusShort~\cite{gumus2024learned}; JPEG-LS~\cite{weinberger2000loco} (\hClsJls) lies above.}
\label{fig:waterfall}
\vspace{2mm}
\includegraphics[width=\linewidth]{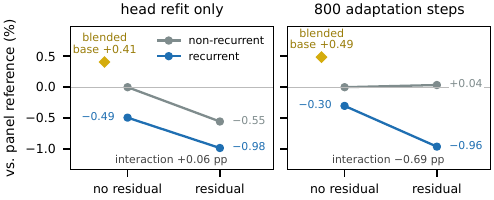}
\caption{Ablation (\nSeven\ images, three seeds, actual bitstreams, no value sets, at the MACs of PerLIC-S): change in bytes against each panel's reference (MED base, non-recurrent state from $h_1$, no residual observation), with head refit only (left) and 800 adaptation steps (right). Diamond: blended base. Values: Supp.~Table~\sref{tab:abl}.}
\label{fig:abl}
\end{figure}

\begin{figure*}[t]
\centering
\includegraphics[width=\textwidth]{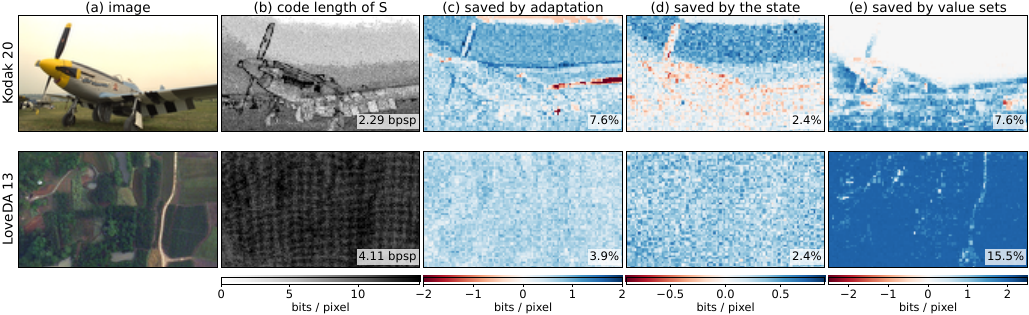}
\caption{Where PerLIC-S saves bits (one seed, 16-bit table; $768{\times}448$ crops, block means; whole-image percent). (b, d, e) Theoretical code length without adaptation; (c) 1600 steps, without minus with adaptation (percent: actual files); (d) non-recurrent minus recurrent state; (e) without minus with value sets; red: cost; one scale per column. Kodak 20: Kodak's top state saving; more: Supp.~Fig.~\sref{fig:mechgrid}.}
\label{fig:mech}
\end{figure*}

\paragraph{What the gain rests on.}
A network fed only causal pixel differences and earlier-color errors is \ledPixGumus\% above \gumusShort; head refit and the classical features bring it to \ledFeatGumus\% below, and adaptation is the largest step (Figure~\ref{fig:waterfall}). Per set, adaptation saves up to \ledAdaMax\% (\ledAdaMaxSet) and the state with the residual observation \ledStateMax\% (\ledStateMaxSet; Supp.~Table~\sref{tab:ledger}). Removing any single part of S as in Table~\ref{tab:ext}, except the composition of the training set, enlarges the files for every seed, by \looMin--\looMax\% (head refit: \looRefit\%; Supp.~Fig.~\sref{fig:loo}).
In separate ablations, the largest single factor is the conditioning on earlier colors: without the decoded errors of red and green at the same pixel, the pixels-only network with non-affine heads is \ledCross\% larger. Making the conditioning affine saves a further \sVtwoGainSEightZeroZero\% for S (800 steps; Supp.~Fig.~\sref{fig:affine}).

\paragraph{What the state adds.}
Figure~\ref{fig:abl} compares four decoders (recurrent state or not, residual observation or not) and the blended base, all at the same MACs.
With the head refit only, the state and the residual observation save about the sum of their single savings (interaction \noaInterSign\ points); after adaptation, the residual alone saves nothing and pays off only with the state (interaction \ablInterSign\ points).
Adaptation saves \adaptGainMin\% to \adaptGainMax\% on every decoder, and it does not replace the state: a non-recurrent state in the full decoder makes the files \ablRmRec\% larger with adapter fitting, for \ablRmRecImgs\ of the \nSeven\ images. In Figure~\ref{fig:mech}, the state saves most in the sky of Kodak 20, adaptation almost everywhere, value sets across LoveDA 13. Over all images, the state saving leans less toward textured pixels than the code length: the smoothest tenth of the pixels of each image holds \sTdStateLoAll\% of it but \sTdBitsLoAll\% of the bits, the most textured tenth \sTdStateHiAll\% and \sTdBitsHiAll\% (theoretical code length without adaptation; Supp.~Sec.~\sref{sec:mechmaps}). Without the state, matching S takes about \sWideExtraMacs\% more MACs (Supp.~Table~\sref{tab:c1}). At about equal MACs, the recursion also beats explicit depth after adaptation: two to four untied hop layers, each reading the eight neighbor states, leave the files \sDepAdBest\% to \sDepAdUFourFb\% larger than the recurrent state (Supp.~Table~\sref{tab:depthad}).

\begin{figure}[t]
\centering
\includegraphics[width=\linewidth]{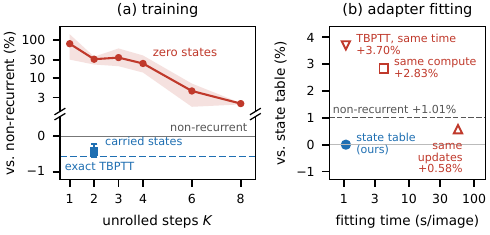}
\caption{Persistent states (three seeds; band and bar: seed range). (a) Training: bitrate under the exact recursion with head refit only, relative to the non-recurrent state, for $K$-step unrolls from zero states, for states carried over the truncation ($K{=}2$, as in our training), and, as a reference, for exact TBPTT. (b) Adapter fitting: fitting time and bitrate relative to the state table, for TBPTT with equal updates, equal time, or the compute of the table in the same implementation (table: compiled; TBPTT: direct implementation; with equal updates, TBPTT takes \tabTbpttSlow$\times$ as long per image); red: TBPTT; dashed: the adapted non-recurrent state.}
\label{fig:buffer}
\end{figure}

\paragraph{States at the truncation.}
Under the exact recursion with head refit only, zero states at the truncation make all \bufTruncRuns\ models ($K=1$ to \bufKmax) larger than a non-recurrent state, by \bufKoneMean\% at $K{=}1$ and \bufKdeepMean\% at $K{=}\bufKmax$ (Figure~\ref{fig:buffer}a; Supp.~Fig.~\sref{fig:zerok}). With the states carried over, S ($K{=}2$) is \bufGain\% smaller than the non-recurrent state. Adapted for 800 steps, the models with $K\le3$ stay larger than the adapted non-recurrent state for every seed, $K{=}4$ ends \bufAdKFourMin\ to $+$\bufAdKFourMax\% from it, and those with $K{=}6$ and \bufKmax\ end \bufAdKOursMin\ to $+$\bufAdKOursMax\% from ours but train up to \bufKdeepTime$\times$ longer.
In adaptation, zeros in place of the table make the files \sZsOneVsTable\% or \sZsTwoVsTable\% larger, and dropping the neighbor-state gradient \tabNoGrad\% (Supp.~Table~\sref{tab:persist}). TBPTT (Figure~\ref{fig:buffer}b) recomputes the states within each segment of the recursion, but an update runs through about \sSeqDepthMean\ sequential wavefronts, one of the table through two levels. With equal updates, TBPTT gives \tabTbpttSame\% larger files at \tabTbpttEager$\times$ the time per update in the same implementation; at the compute of the table, it gives files \tabTbpeVsNonrec\% larger than the adapted non-recurrent state (\sTbfVsNr\% with the \sTbfUpd\ updates that the table costs on large images; also with doubled learning rate or compute, Supp.~Sec.~\sref{sec:persist}). At matched fitting compute, recurrence thus pays off only with the table.

\paragraph{Value sets.}
At 1600 steps, the encoder keeps value sets on all Kodak, RS19, and Histo images and on \vsOnLoveda\ LoveDA images (\vsCoarseOnLoveDA\ with coarse keys). There the colors take few combinations (on Kodak because of the Photo CD color transform; Supp.~Sec.~\sref{sec:lattice}), and the sets make Kodak \vsKodakGain\% and Histo \hiVsGain\% smaller; on DIV2K, CLIC, and Doc, always-on sets would cost up to \vsOtherCostMax\% (Supp.~Table~\sref{tab:vs}). Over all images, they save \vsAllGain\% (\vsAlwaysGain\% always on; Supp.~Table~\sref{tab:vscfg}). This saving barely depends on texture (theoretical code length, sets always on; Supp.~Sec.~\sref{sec:mechmaps}). The saving splits into the head's redundancy on set membership and the regret of the adaptive escape against the best constant escape per context in hindsight, which stays within \vsDecRMax\ points per set (Supp.~Sec.~\sref{sec:vsdecomp}), so almost all of the saving is the redundancy of the head. On CALLIC, a mixture of five logistics, they save \cvsRs\% (RS19) to \cvsKodak\% (Kodak) on these sets. In S, a mixture of four logistics saves \sMixAll\% (800 steps), nearly as much as the sets, but decodes \sMixDecEight\% slower (8 threads; Supp.~Sec.~\sref{sec:mix}); it adds work on every image, the sets only where the switch keeps them.

\section{Conclusion and limitations}
\label{sec:conclusion}
A small integer decoder, adapted to each image by a GPU encoder, keeps much of the rate advantage of learned compression while decoding bit-exactly on a CPU, for S at about the speed of JPEG XL on eight AVX-512 cores. Its recurrent state reuses earlier computation and pays off once its short unrolls start not from zeros but from stored states, which a persistent state table supplies cheaply in adaptation; decoder-side value sets reclaim the probability that its head, with few peaks, spreads over the gaps.

\paragraph{Limitations.}
We evaluate 8-bit RGB images only, and adapted encoding needs a GPU. S is slower than JPEG XL with one thread (Supp.~Table~\sref{tab:t21}), with eight threads on an Apple M4 Pro (\sMproRatioEight$\times$; Supp.~Sec.~\sref{sec:xdev}), and with AVX2 alone (\sIsaAvxOne$\times$ its AVX-512 time; Supp.~Table~\sref{tab:isa}). It also needs \sRssS\ MB of memory against \sRssJxl\ MB for JPEG XL.

\section*{LLM Usage Statement}
We used Claude (Anthropic) to extend our code, run experiments, and draft the paper. The authors wrote the base code, designed the study, checked every result, and take full responsibility.

{\small
\bibliographystyle{ieeenat_fullname}
\bibliography{refs}
}

\end{document}